\PassOptionsToPackage{colorlinks,linkcolor=red,anchorcolor=blue,citecolor=green}{hyperref}
\documentclass[11pt]{article}

\usepackage[preprint]{acl}

\usepackage{times}
\usepackage{latexsym}

\usepackage[T1]{fontenc}

\usepackage[utf8]{inputenc}

\usepackage{microtype}

\usepackage{inconsolata}

\usepackage{graphicx}

\usepackage[linesnumbered,ruled,longend]{algorithm2e}
\usepackage{algorithmic}

\usepackage{adjustbox}
\usepackage{multirow}
\usepackage{newfloat}
\usepackage{listings}
\usepackage{enumitem}
\usepackage{amsmath}
\usepackage{amssymb}
\usepackage{amsthm}
\usepackage{caption}
\usepackage{colortbl}
\usepackage{xcolor}
\usepackage{booktabs}
\usepackage{color}
\usepackage{newfloat}
\usepackage{listings}
\usepackage{tabularx}  
\usepackage{array}     
\usepackage{adjustbox} 
\usepackage[colorlinks,
            linkcolor=red,
            anchorcolor=blue,
            citecolor=green
            ]{hyperref}
\usepackage[listings]{tcolorbox}
\usepackage[table]{xcolor}
\newcommand{\token}[1]{\texttt{\small\textless#1\textgreater}}
\definecolor{lightgreen}{rgb}{0.85,1.0,0.85}

\newcommand{\colorEvidence}[1]{\textcolor[rgb]{0.1,0.3,0.5}{\textbf{#1}}}

\title{SE-Search: Self-Evolving Search Agent via Memory and Dense Reward}

\author{
 \textbf{Jian Li\textsuperscript{1,2}\textsuperscript{*}\textsuperscript{$\dag$}},
 \textbf{Yizhang Jin\textsuperscript{2}},
 \textbf{Dongqi Liu\textsuperscript{2}},
 \textbf{Hang Ding\textsuperscript{2}},
 \textbf{Jiafu Wu\textsuperscript{2}},
 \textbf{Dongsheng Chen\textsuperscript{2}},
 \\
 \textbf{Yunhang Shen\textsuperscript{2}},
 \textbf{Yulei Qin \textsuperscript{2}},
 \textbf{Ying Tai\textsuperscript{1}},
 \textbf{Chengjie Wang\textsuperscript{2}},
 \textbf{Xiaotong Yuan\textsuperscript{1}\textsuperscript{$\dag$}},
 \textbf{Yabiao Wang\textsuperscript{2}\textsuperscript{*}},
\\
 \textsuperscript{1}Nanjing University,
 \textsuperscript{2}Tencent YoutuLab,
\\
 \small{
   $^*$ Project leader. $^\dagger$ Corresponding author.
 }
}

\begin{document}
\maketitle
\begin{abstract}
Retrieval augmented generation (RAG) reduces hallucinations and factual errors in large language models (LLMs) by conditioning generation on retrieved external knowledge. Recent search agents further cast RAG as an autonomous, multi-turn information-seeking process. However, existing methods often accumulate irrelevant or noisy documents and rely on sparse reinforcement learning signals. We propose \textbf{S}elf-\textbf{E}volving \textbf{Search}, a Self-Evolving Search agent that improves online search behavior through three components, memory purification, atomic query training, and dense rewards. SE-Search follows a \textit{Think-Search-Memorize} strategy that retains salient evidence while filtering irrelevant content. Atomic query training promotes shorter and more diverse queries, improving evidence acquisition. Dense rewards provide fine-grained feedback that speeds training. Experiments on single-hop and multi-hop question answering benchmarks show that \texttt{SE-Search-3B} outperforms strong baselines, yielding a $10.8$ point absolute improvement and a $33.8\%$ relative gain over Search-R1.\footnote{We will make the code and model weights publicly available upon acceptance.}
\end{abstract}

\section{Introduction}

Searching for information~\cite{case2016lookinginfo} is a common activity in daily life. Search engines such as Google and Bing return ranked web pages for a user query, enabling fast access to relevant content. As queries become more complex and users expect higher precision, traditional information retrieval techniques~\cite{kobayashi2000ir} may fail to capture subtle intent and return results that do not match the user’s context.

Large language models (LLMs)~\cite{zhao2023surveyllm} have shown strong capabilities in language understanding, reasoning, and information integration. However, they do not reliably access up-to-date external knowledge and may produce factual errors. Retrieval-augmented generation (RAG)~\cite{lewis2020rag} reduces these problems by conditioning generation on retrieved external passages. Yet fixed RAG pipelines limit an LLM's ability to decide when to search and what to search for, which reduces flexibility when interacting with real-world applications.

Recent work~\cite{li2025aissurvey} presents LLM-based agents~\cite{guo2024agentsurvey} that interpret user intent, plan search strategies, conduct multi-turn search actions, and accumulate retrieved evidence. These systems combine the strengths of LLMs and search engines and are often referred to as search agents. Representative methods such as Search-R1~\cite{jin2025searchr1} use reinforcement learning (RL)~\cite{kaelbling1996rl,shao2024deepseekmath} with rule-based rewards to train agents to interact with a search tool over a large corpus.

\begin{figure*}[h]
	\centering
	\includegraphics[width=0.95\linewidth]{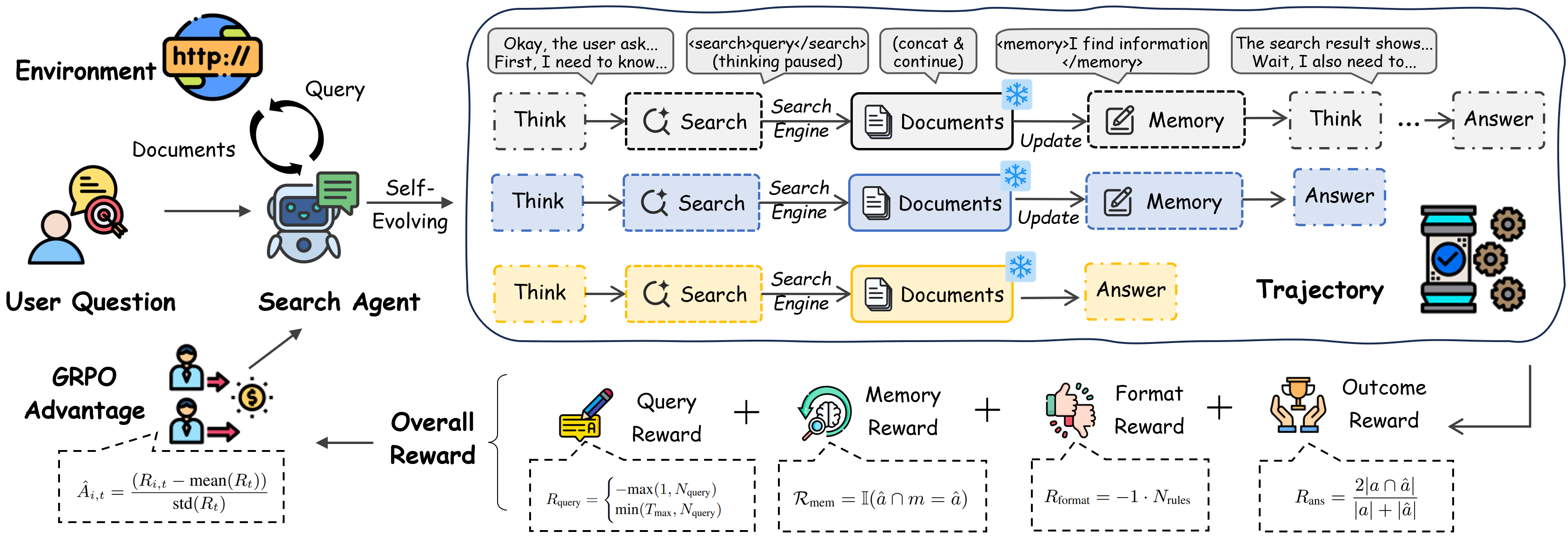}
	\caption{Training scheme of SE-Search. For each question, the search agent generates diverse trajectories comprising the steps \textit{think}, \textit{search}, \textit{memorize}, and \textit{answer}. These trajectories are optimized using the GRPO algorithm \cite{shao2024deepseekmath} and four carefully designed rewards: Query, Format, Memory, and Outcome.}
	\label{fig_framework}
    \vspace{-1em}
\end{figure*}

Despite promising results, existing search agent methods face three key challenges. \textbf{Noisy search results:} Agents typically retrieve top-$K$ documents after issuing a query, and many are irrelevant or noisy, which provides limited support for answering user questions. MAIN-RAG~\cite{chang2025mainrag} reduces noise through multi-LLM filtering and scoring, but it operates in a training-free RAG setting. \textbf{Limited search diversity and underexplored search frequency:} Prior methods~\cite{shi2025autorefine} often generate similar queries across search steps, which limits exploration of diverse and informative evidence. Although $O^2$-Searcher~\cite{mei20252o2searcher} introduces a query diversity reward based on embedding similarity, the additional encoding step reduces efficiency. \textbf{Sparse evolutionary feedback signals:} Search-R1 provides supervision at the final answer level and does not explicitly reward query formulation, formatting, or appropriate search frequency. As a result, agents may produce overly long queries regardless of question complexity, which limits effectiveness on multi-hop questions. TooRL~\cite{qian2025toolrl} proposes richer rewards, but it targets general tool use rather than search-specific behavior.

Inspired by Darwin’s theory of evolution~\cite{ruse1975darwin,ang2507self-evolvingsurvey}, which emphasizes adaptation to changing environments, we focus on how reward and feedback signals guide agents on complex real-world queries. These signals do not provide facts directly, but they shape how an agent discovers, processes, and uses external information. We therefore propose \texttt{SE-Search}, a self-evolving search agent that improves an LLM’s autonomous search behavior. SE-Search adopts a \textit{memorize-after-search} paradigm and extracts useful evidence with a \textbf{Memory Purification} template to reduce noisy retrievals. Unlike DeepAgent~\cite{li2025deepagent}, which relies on auxiliary LLMs for memory management, SE-Search uses the agent’s self-memory. To promote diverse search behaviors and appropriate search frequency, we introduce an \textbf{Atomic Query} strategy motivated by Atom-Searcher~\cite{deng2025atomsearcher}, which guides the agent to generate multiple distinct atomic queries and perform multi-step searches. Finally, we design \textbf{Dense Rewards} composed of four components, Query, Memory, Outcome, and Format, to provide fine-grained RL feedback, improve behavioral discipline, and stabilize training.

We summarize the contributions as follows. 

\begin{itemize}
\vspace{-5pt}
\item We propose SE-Search, a self-evolving search agent that improves adaptability to complex, real-world questions. 
\vspace{-5pt}
\item We steer the agent's evolution by introducing three mechanisms: Memory Purification, Atomic Query, and Dense Rewards.
\vspace{-5pt}
\item We demonstrate the effectiveness and generalizability of SE-Search on seven diverse and challenging QA benchmarks.
\end{itemize}

\section{Problem Formulation}
\label{sec_problem_formulation}

We model question answering task as

\begin{small}
\begin{equation}
a = \pi_{\theta}(Q)
\end{equation}
\end{small}

where $Q$ is the user question, $a$ is the generated answer, and $\pi_{\theta}$ denotes the LLM. 

In retrieval augmented generation, the system retrieves knowledge $k$ from an external corpus $\mathcal{D}$ using a retriever $\mathcal{R}(\cdot)$ and then generates the answer conditioned on $Q$ and $k$.

\begin{small}
\begin{equation}
a = \pi_{\theta}(Q \mid k), \quad k = \mathcal{R}(Q \mid \mathcal{D})
\end{equation}
\end{small}

However, this fixed workflow is often insufficient for complex knowledge discovery tasks that require iterative reasoning and adaptive retrieval. Recent advances in search agents support a more dynamic paradigm in which retrieval is interleaved with reasoning, and the model output $y$ may include intermediate reasoning steps $r$ and the final answer $a$.

\begin{small}
\begin{equation}
a = \pi_{\theta}\left(Q \mid  \{r_t \otimes k_t\}_{t=1}^{T}\right),  
k_t = \mathcal{R}(q_t \mid \mathcal{D}),  q_t \in r_t
\end{equation}
\end{small}

where $q_t$ denotes subqueries generated during reasoning to retrieve intermediate knowledge $k_t$, and the operator $\otimes$ indicates that retrieved knowledge is integrated into the reasoning trajectory. Ignoring the intermediate trajectory, we employ supervised fine-tuning (SFT) to optimize only the final answer using the next-token prediction loss, which is defined as follows. 

\begin{small}
\begin{equation}
\max_{\theta} \mathbb{E}\big[\log \pi_{\theta}(\hat{a})\big]
\end{equation}
\end{small}

where $\mathbb{E}$ denotes expectation. This objective maximizes the expected log likelihood that $\pi_{\theta}$ assigns to the correct answer $\hat{a}$.

\section{Approach}

In this section, we present \texttt{SE-Search}, a self-evolving search agent that improves autonomous search behavior. We first formalize the search agent setting and its optimization objective. We then describe three components: memory purification, atomic query generation, and dense rewards.

\subsection{Self-Evolving Search Agent}
\label{sec_search_agent}

Figure~\ref{fig_framework} illustrates the optimization scheme of SE-Search. Given a user question, the agent interacts with a web environment through a search tool to generate multiple trajectories $\tau$. Each trajectory is represented as $\tau = (\tau_1, \tau_2, \dots, \tau_T)$, where step $t$ is denoted as $\tau_t = (a_t, c_t)$. The action $a_t$ is selected from $\{\textcolor{blue}{\token{think}}, \textcolor{green}{\token{search}}, \textcolor{orange}{\token{documents}}, \textcolor{purple}{\token{memory}}, \textcolor{red}{\token{answer}}\}$, and the content $c_t$ corresponds to a reasoning thought, a search query, retrieved passages, a distilled summary, or a final answer. The agent aims to gather relevant evidence across steps and produce an accurate final answer.

Formally, we optimize the following objective.

\begin{small}
\begin{equation}
\max_{\theta} \mathbb{E}\big[\log \pi_{\theta}(\hat{a})  + \alpha \cdot \mathcal{C}( \cup_{t=1}^{T} k_t, \hat{K} ) \big]
\label{eq_inf}
\end{equation}
\end{small}

where $\mathcal{C}(\cdot, \cdot)$ measures the coverage of retrieved documents relative to the expected knowledge $\hat{K}$, $T$ denotes the number of retrieval steps, and $\alpha \in (0, 1)$ is a weighting factor. Search-R1~\cite{jin2025searchr1} shows that loss masking can exclude retrieved tokens from gradient optimization. Following this insight, we approximate retrieval using the search queries and memory contents and rewrite the objective accordingly.

\begin{small}
\begin{equation}
\max_{\theta} \mathbb{E}\big[\log \pi_{\theta}(\hat{a})  + \alpha \cdot \sum_{t}\log \pi_{\theta}(\hat{m_t}) + \gamma \cdot \sum_{t}\log \pi_{\theta}(\hat{q_t}) \big]
\label{eq_inf_approx}
\end{equation}
\end{small}

where $\gamma$ is an additional weighting factor. This objective accounts for three aspects: search queries, memory, and final answers. Due to the lack of ground truth trajectory, supervised fine-tuning is not feasible. We adopt a post-training reinforcement learning framework and use a rule-based reward function to optimize trajectory generation.

\begin{figure}[t]
    \centering
    \small
    \vspace{-.5em}
    \begin{tcolorbox}[colback=gray!5!white, colframe=black!15, width=\linewidth, boxrule=0.5pt, arc=2mm]
   You are a capable reasoning assistant, able to perform multiple search calls and memory purification to answer questions. You must reason through the available information using \textcolor{blue}{\token{think}} and \textcolor{blue}{\token{/think}}. If you lack knowledge, you can call a search engine using \textcolor{green}{\token{search}} query \textcolor{green}{\token{/search}} and it will return the top three results between \textcolor{orange}{\token{documents}} and \textcolor{orange}{\token{/documents}}.
    After each search, extract useful information from these documents and supplement or revise your memory between \textcolor{purple}{\token{memory}} and \textcolor{purple}{\token{/memory}}.
    You may send multiple search requests if needed. Do not repeat search queries. Once you have sufficient information, provide a concise final answer using \textcolor{red}{\token{answer}} and \textcolor{red}{\token{/answer}}. For example, <answer> Donald Trump </answer>. Question \{\textbf{question}\} 
    \end{tcolorbox}
    \vspace{-.5em}
    \caption{Prompt template for SE-Search.}
    \label{fig_prompt_template}
\end{figure}

\subsection{Memory Purification.}
\label{sec_mp}

Prior methods forward-retrieve documents to the LLM without filtering, which forces the model to reason over irrelevant and noisy content. Inspired by Memory-R1~\cite{yan2025memoryr1}, we introduce memory purification to filter and consolidate relevant facts from retrieval documents and store key information using the special token \textcolor{purple}{\token{memory}}. Figure~\ref{fig_prompt_template} shows the employed prompt require LLM to supplement or revise previous memory knowledge by incorporating retrieved knowledge $k$ into latest memory $m_t$. The update process is modeled as:

\begin{small}
\begin{equation}
m_t = \pi_{\theta}(m_{t-1} \mid  k_t)
\end{equation}
\end{small}

This strategy filters and integrates only useful information, thereby maintaining coherent and evolving knowledge. To explicitly encourage selective extraction from noisy retrieved documents, we define a memory reward. Following AutoRefine~\cite{shi2025autorefine}, the memory reward is measured using Cover Exact Match (CEM) between the memory contents $m$ and the correct answers $\hat{a}$.

\begin{small}
\begin{equation}
\mathcal{R}_{\text{mem}} = \mathbb{I}(\hat{a} \cap m = \hat{a} )
\end{equation}
\end{small}

where $\mathbb{I}(\cdot)$ is the indicator function.

\subsection{Atomic Query}

In Algorithm~\ref{alg_algorithm}, we propose an atomic query counting method that constrains query length and enforces diversity across queries. Based on the number of valid atomic queries $N_{query}$ in a trajectory, we define a query reward that guides when and how the agent invokes the search tool.

\begin{small}
\begin{equation}
    R_{\text{query}} = 
    \begin{cases}
     -\text{max}(1, N_{\text{query}}) & \text{if the answer is correct}\\
     \text{min}(T_{\text{max}}, N_{\text{query}}) & \text{if the answer is incorrect}\\
    \end{cases}
\end{equation}
\end{small}

where $T_{\text{max}}$ is the maximum allowed number of search action turns, and $N_{\text{query}}$ denotes the number of valid search queries in the trajectory. This reward serves two purposes. It encourages the agent to produce diverse queries that improve evidence coverage. It also discourages unnecessary search when the final answer is correct and encourages more search when the final answer is incorrect.

\subsection{Dense Rewards}
Apart from the query and memory rewards, the dense reward includes two additional components: an outcome-based reward that directly measures the correctness of the model's final answer and a format-based reward that encourages adherence to the prescribed reasoning structure. Rather than using a binary exact match signal, we treat the predicted answer and the ground-truth answer as sets and compute the F1 score between the model's predicted answer $a$ and the ground-truth answer $\hat{a}$.

\begin{small}
\begin{equation}
\label{eq_r_ans}
    R_{\text{ans}} =\frac{2|a\cap \hat{a}|}{|a|+|\hat{a}|}
\end{equation}
\end{small}

We define the format reward through three rules that penalize unreasonable trajectories to prevent mode collapse and degenerate behavior. These rules include cases where the trajectory exceeds the predefined maximum number of search action turns $T_{\text{max}}$, contains invalid actions, or includes unmatched or incorrectly ordered special tokens. Let $N_{\text{rules}}$ denote the total number of format violations in the trajectory, and we define

\begin{small}
\begin{equation}
\label{eq_r_format}
    R_{\text{format}} = -1 \cdot N_{\text{rules}}
\end{equation}
\end{small}

Combining all dense reward components, we obtain the overall dense reward

\begin{small}
\begin{equation}
\label{eq_r_overall}
R_{\text{Dense}} = R_{\text{ans}} + \alpha \cdot R_{\text{mem}} + \gamma \cdot \mu \cdot R_{\text{query}} + \gamma \cdot R_{\text{format}} 
\end{equation}
\end{small}

where $R_{\text{mem}}$ and $R_{\text{query}}$ denote the memory and query rewards defined earlier, and $\alpha$ and $\gamma$ are weighting coefficients that balance the reward terms. To improve RL stability, we gradually reduce the influence of the query reward during training by multiplying it with a time-dependent decay factor. The cosine decay schedule is defined as

\begin{small}
\begin{equation}
\begin{aligned}
    \mu = 
    \begin{cases}
     \frac{1}{2} (cos( \frac{t_{\text{iter}}}{t_{\text{decay}}} \pi )+1), &  t_{\text{iter}} \le t_{\text{decay}} \\
    0, & t_{\text{iter}} > t_{\text{decay}}\\
    \end{cases}
\end{aligned}
\end{equation}
\end{small}

where $t_{\text{iter}}$ is the current training iteration and $t_{\text{decay}}$ is the number of decay steps.

\setlength{\textfloatsep}{8pt}
\begin{algorithm}[t]
	\caption{Atomic Query Counting}
	\label{alg_algorithm}
	\KwIn{ Search query set $\mathcal{Q}$ from a trajectory, initialize valid query set $\mathcal{Q}_v \leftarrow \varnothing$, length bounds $L_{\min}, L_{\max}$, similarity threshold $T$. }
	\For{$q_i \in \mathcal{Q}$}{
        \If{ $L_{\min} \le |q_i| \le L_{\max}$}{
            \For{$q_j \in \mathcal{Q}_v$}{
                 Compute matching segment lengths $\ell_k$ between $q_i$ and $q_j$ via SequenceMatch;
                \begin{small}
                \begin{equation}
                \text{ratio}(q_i,q_j) \;=\; \frac{2 \sum_{k} \ell_k}{|q_i| + |q_j|}
                \label{ratio}
                \end{equation}
                \end{small}
                \If{ $\text{ratio}(q_i,q_j) \le T$} { 
                    $\mathcal{Q}_v = \mathcal{Q}_v \cup \{q_i \}$; \\
                }
            }
        }
    }
    \textbf{Return} $N_{query} \leftarrow |\mathcal{Q}_v|$ ;
\end{algorithm}

\label{sec_exp_main}
\begin{table*}[t]
    \centering
    \resizebox{\linewidth}{!}{ 
    \begin{tabular}{lcccccccccc}
        \toprule
        & \multicolumn{4}{c}{Single-Hop QA (EM)} & \multicolumn{5}{c}{Multi-Hop QA (EM)}       & \multicolumn{1}{l}{QA (EM)} \\
        \cmidrule(lr){2-5} \cmidrule(lr){6-10} \cmidrule(lr){11-11}
        Methods            & NQ      & TriviaQA   & PopQA  & Avg.   & HotpotQA  & 2Wiki & Musique & Bamboogle & Avg.  & Avg. \\
        \midrule
        \multicolumn{9}{l}{{w/o Retrieval}} \\
        \quad Direct Generation     & 0.106 & 0.288 & 0.108 & 0.167          & 0.149 & 0.244 & 0.020 & 0.024 & 0.109       & 0.134 \\
        \quad SFT                  & 0.249   & 0.292  & 0.104 &0.215  & 0.186    & 0.248 & 0.044   & 0.112  & 0.148   & 0.176 \\
        \quad R1-Instruct \cite{guo2025deepseekr1}          & 0.210   & 0.449      & 0.171  & 0.277 & 0.208    & 0.275 & 0.060   & 0.192  & 0.184   & 0.224 \\
        \quad R1-Base \cite{guo2025deepseekr1}              & 0.226   & 0.455      & 0.173 & 0.285  & 0.201    & 0.268 & 0.055   & 0.224 &0.187     & 0.229 \\
        \midrule
        \multicolumn{9}{l}{{Workflow w/ Retrieval}} \\
        \quad Naive RAG \cite{lewis2020rag}                  & 0.348   & 0.544      & 0.387 & 0.426  & 0.255    & 0.226 & 0.047   & 0.080   & 0.152  & 0.270 \\
        \quad IRCoT \cite{trivedi2022ircot} & 0.111   & 0.312      & 0.200  & 0.208 & 0.164    & 0.171 & 0.067   & 0.240  & 0.161   & 0.181 \\
        \midrule
        \multicolumn{9}{l}{{Agent w/ Retrieval}} \\
        \quad Search-o1 \cite{li2025search-o1}                 & 0.238   & 0.472      & 0.262 & 0.324  & 0.221    & 0.218 & 0.054   & 0.320   & 0.203  & 0.255 \\
        \quad Search-R1-Instruct \cite{jin2025searchr1}       & 0.397   & 0.565      & 0.391 & 0.451  & 0.331    & 0.310 & 0.124   & 0.232  & 0.249   & 0.336 \\
        \quad Search-R1-Base \cite{jin2025searchr1}           & 0.421   & 0.583      & 0.413  & 0.472 & 0.297    & 0.274 & 0.066   & 0.128  & 0.191   & 0.312 \\
        \quad ReSearch-Instruct \cite{chen2025ReSearch}         & 0.365                & 0.571                & 0.395 & 0.444 & 0.351 & 0.272 & 0.095 & 0.266 & 0.246 & 0.331 \\
        \quad ReSearch-Base \cite{chen2025ReSearch}             & 0.427          & 0.597          & 0.430 & 0.485 & 0.305 & 0.272 & 0.074 & 0.128 & 0.195 & 0.319 \\
        \quad ZeroSearch-Base \cite{sun2025zerosearch}             & 0.430          & 0.616          & 0.414 & 0.487 & 0.338 & 0.346 & 0.130 & 0.139 & 0.238 & 0.345 \\
        \quad StepSearch-Base \cite{wang2025stepsearch}     & -    & -          & - & - & 0.329 & 0.339 & 0.181 & 0.328 & 0.294 & - \\
        \quad $O^2$-Searcher
        \cite{mei20252o2searcher}       & 0.444   & 0.597      & 0.429  & 0.490 & 0.388    & 0.374 & 0.160   & 0.344 & 0.317    & 0.391 \\
        \quad AutoRefine-Instruct \cite{shi2025autorefine}             & 0.436 & 0.597 & 0.447 & 0.493 & 0.404 & 0.380 & 0.169 & 0.336 & 0.322 & 0.396 \\
        \quad AutoRefine-Base \cite{shi2025autorefine}                   & 0.467 & 0.620 & 0.450 & \textbf{0.512}  & 0.405 & 0.393 & 0.157 & 0.344 & 0.325 & 0.405 \\        
        \quad InForage
        \cite{qian2025inforage}       & 0.421   & 0.597      & \textbf{0.452} & 0.490  & 0.409    & \textbf{0.428} & 0.172   & 0.360   & 0.342  & 0.405 \\
        \quad CriticSearch \cite{zhang2025criticsearch}       & -    & -          & - & - & 0.414 & 0.409 & 0.180 & 0.368 & 0.343 & - \\
        \midrule
        \rowcolor{purple!10}
        \quad \textbf{SE-Search-3B} (Ours)                  & \textbf{0.475} & \textbf{0.624} & 0.423 & 0.507 & \textbf{0.450} & 0.361 & \textbf{0.183} & \textbf{0.424} & \textbf{0.355} & \textbf{0.420} \\
        \bottomrule
        \end{tabular}
    }
    \vspace{-0.5em}
    \caption{
    Accuracy comparison of SE-Search-3B against baseline methods using Qwen2.5-3B \cite{qwen2025qwen25technicalreport} across multiple QA benchmarks. \textbf{Bold} indicates best results, higher values denote better performance.
    }
    \label{tab_exp_main}
\vspace{-1em}
\end{table*}

\subsection{Agentic Reinforcement Learning}
\label{sec_rl}

To avoid a separate value estimator, we adopt Group Relative Policy Optimization (GRPO) \cite{shao2024deepseekmath} as our agentic reinforcement learning framework. For a question $Q$ from the dataset $D$, the policy model $\pi_\theta$ samples an interleaved sequence ${y_{0}, k_{0}, y_{1}, k_{1}, \ldots, y_{T}, k_{T}}$. Following Search-R1 ~\cite{jin2025searchr1}, external retrieval tokens $k$ are masked during loss computation, so the objective depends only on the model outputs $y$.

\begin{small}
\begin{equation}
    \label{eq_grpo}
    \begin{aligned}
     \mathcal{L}_{\text{GRPO}}(\theta) = 
     \mathbb{E}_{Q \sim \mathcal{D}, y_i\sim \pi_{\theta}}
    \Big[
    \frac{1}{G} \sum_{i=1}^{G} 
    \frac{1}{|y_i|} \sum_{t=1}^{|y_i|}
    \min (r_{i,t} \hat{A}_{i,t}, \quad \quad 
    \\ 
    \text{clip} \left(r_{i,t}, 1-\epsilon, 1+\epsilon\right) \hat{A}_{i,t})
    -  \beta \mathcal{KL} \left[ \pi_{\theta} \, \| \, \pi_{\text{ref}} \right]
    \Big]
    \end{aligned}
\end{equation}
\end{small}

where $\pi_{\theta}$ denotes the current actor and $\pi_{\text{ref}}$ denotes a fixed reference policy. The probability ratio is computed using the previous policy $\pi_{\theta_{\text{old}}}$.

\begin{small}
\begin{equation}
r_{i,t} = \frac{ \pi_{\theta}(y_{i,t} \mid Q, y_{i,<t}) }
     { \pi_{\theta_{\text{old}}}(y_{i,t} \mid Q, y_{i,<t}) }
\end{equation}
\end{small}

For each question, we use a group of $G$ sampled outputs and normalize their rewards to estimate advantages.

\begin{small}
\begin{equation}
\hat{A}_{i,t} =\frac{ (R_{i,t} - \text{mean}(R_t)) }  {\text{std}(R_t)} 
\end{equation}
\end{small}

where $R$ is the overall dense reward in Eq.~\ref{eq_r_overall}.

\section{Experiments}
\label{sec_exp}

\subsection{Experiment Settings}
\label{sec_exp_setup}

\paragraph{Benchmarks and Datasets.}

We evaluate SE-Search on seven question answering benchmarks that require single-hop or multi-hop retrieval. The single-hop datasets include Natural Questions (NQ) \cite{kwiatkowski2019NQ}, TriviaQA \cite{joshi2017triviaqa}, and PopQA \cite{mallen2022popqa}. The multi-hop datasets include HotpotQA \cite{yang2018hotpotqa}, 2WikiMultihopQA \cite{ho2020wikimultihopqa}, Musique \cite{trivedi2022musique}, and Bamboogle \cite{press2022bamboogle}. We use exact match (EM) for all datasets. Following Search-R1 \cite{jin2025searchr1}, we train SE-Search on a combined training set of NQ and HotpotQA.

\paragraph{Baselines.}
We compare SE-Search with three classes of methods.
(1) Methods without retrieval include direct LLM generation, supervised fine-tuning (SFT), and R1-style training \cite{guo2025deepseekr1};
(2) Workflows with retrieval include naive RAG that retrieves only from the input question and IRCoT \cite{trivedi2022ircot}, which interleaves retrieval with chain-of-thought;
(3) Agents with retrieval include retrieval-augmented methods such as Search-o1 \cite{li2025search-o1}, Search-R1 \cite{jin2025searchr1}, ReSearch \cite{chen2025ReSearch}, ZeroSearch \cite{sun2025zerosearch}, StepSearch \cite{wang2025stepsearch}, $O^2$-Searcher \cite{mei20252o2searcher}, InForage \cite{qian2025inforage}, CriticSearch \cite{zhang2025criticsearch}, and AutoRefine \cite{shi2025autorefine}.

\paragraph{Implementation Details.}
To emulate a standard search setting, we use the external corpus \cite{karpukhin2020wikipediadump} used by Search-R1 and adopt E5-base-v2 \cite{wang2022e5} as the retriever.
By default, the retriever returns the top three documents per query, and the backbone LLM is Qwen2.5-3B~\cite{qwen2025qwen25technicalreport}. The query length bounds $L_{\min}$ and $L_{\max}$ are set to 20 and 120 characters. The similarity threshold is $T = 0.3$. The maximum number of search turns is $T_{\text{max}} = 5$, and the decay steps $t_{\text{decay}}$ are set to 150. The reward weights are $\alpha = 0.1$ and $\gamma = 0.01$.

\begin{table*}[t]
    \centering
    \small
    \resizebox{0.95\linewidth}{!}{
    \begin{tabular}{c|cccccccc}
        \toprule
        \multirow{3}{*}{ Method } & \multicolumn{3}{c}{Single-Hop QA} & \multicolumn{4}{c}{Multi-Hop QA}       & \multicolumn{1}{l}{} \\
        \cmidrule(lr){2-4} \cmidrule(lr){5-8}
        & NQ    & TriviaQA   & PopQA   & HotpotQA & 2Wiki & Musique & Bamboogle & Avg.                 \\
        \midrule
        \rowcolor{gray!10} Search-R1 &  0.424 & 0.630   & 0.471   & 0.346 & 0.296 & 0.070 & 0.169 & 0.344 \\
        \hline
        \rowcolor{gray!10} Search-R1   &  0.426 & 0.610  & 0.436  & 0.377 & 0.315 & 0.120 & 0.260 & 0.363 \\ 
        w/ MP &   \textcolor{purple}{$+0.002$} & \textcolor{cyan}{$-0.020$} & \textcolor{cyan}{$-0.035$} & \textcolor{purple}{$+0.031$} & \textcolor{purple}{$+0.019$} & \textcolor{purple}{$+0.050$} & \textcolor{purple}{$+0.091$} & \textcolor{purple}{$+0.019$} \\
         & $\uparrow0.47\%$ & $\downarrow3.17\%$ & $\downarrow7.43\%$ & $\uparrow8.96\%$ & $\uparrow6.42\%$ & $\uparrow71.43\%$ & $\uparrow53.85\%$ & $\uparrow5.52\%$ \\
         \hline
         \rowcolor{gray!10}  Search-R1  &  \textbf{0.442} & 0.609  & 0.427 & 0.381 & 0.365 & 0.163 & \textbf{0.371} & 0.394 \\
        w/ MP, AQ &   \textcolor{purple}{$+0.016$} & \textcolor{cyan}{$-0.001$} & \textcolor{cyan}{$-0.009$} & \textcolor{purple}{$+0.004$} & \textcolor{purple}{$+0.050$} & \textcolor{purple}{$+0.043$} & \textcolor{purple}{$+0.111$} & \textcolor{purple}{$+0.031$} \\
         & $\uparrow3.76\%$ & $\downarrow0.16\%$ & $\downarrow2.06\%$ & $\uparrow1.06\%$ & $\uparrow15.87\%$ & $\uparrow35.83\%$ & $\uparrow42.69\%$ & $\uparrow8.54\%$ \\
          \hline
        \rowcolor{gray!10}  Search-R1  & 0.429 & \textbf{0.661} & \textbf{0.467} & \textbf{0.436} & \textbf{0.374} & \textbf{0.177} & 0.328 & \textbf{0.410} \\
        w/ MP, AQ, DR &   \textcolor{cyan}{$-0.013$} & \textcolor{purple}{$+0.052$} & \textcolor{purple}{$+0.040$} & \textcolor{purple}{$+0.055$} & \textcolor{purple}{$+0.009$} & \textcolor{purple}{$+0.014$} & \textcolor{cyan}{$-0.043$} & \textcolor{purple}{$+0.016$} \\
         (SE-Search) & $\downarrow2.94\%$ & $\uparrow8.55\%$ & $\uparrow9.37\%$ & $\uparrow14.44\%$ & $\uparrow2.47\%$ & $\uparrow8.59\%$ & $\downarrow11.59\%$ & $\uparrow4.06\%$ \\
        \bottomrule
        \end{tabular}
    }
    \vspace{-0.5em}
    \caption{
    Ablation results showing the effect of individual SE-Search components. The first row reports the baseline method. Subsequent rows show performance after adding Memory Purification, Atomic Query, and Dense Rewards.
    }
    \label{tab_exp_ablation}
\vspace{-1.5em}
\end{table*}

\subsection{Main Performance}
\label{sec_exp_main}

Table~\ref{tab_exp_main} presents the main experimental results comparing SE-Search with baseline methods. The {Avg.} column reports the average accuracy. From these results, we make the following key observations.

\textbf{SE-Search outperforms other methods.} Under the same retriever, corpus, training data, and Qwen2.5-3B setting, SE-Search consistently outperforms Search-R1 and recent methods, including InForage and AutoRefine, across seven benchmarks covering both single-hop and multi-hop QA. SE-Search achieves an average EM accuracy of $0.420$, demonstrating that the proposed memory purification mechanism and dense reward design effectively enhance the self-evolution capability of search agents during information seeking.

\textbf{SE-Search shows particularly strong gains on multi-hop QA benchmarks.} The performance improvements are more evident on complex multi-hop QA tasks. For instance, compared with AutoRefine, SE-Search improves HotpotQA by $4.5$ percentage points, corresponding to an $11.1\%$ relative gain, and improves Bamboogle by $8$ percentage points, corresponding to a $23.2\%$ relative gain. These improvements can be attributed to two key design choices. Atomic queries decompose the original question into multiple subqueries, enabling more effective retrieval across steps, while memory purification reduces noise accumulation during multi-step search and reasoning.

\subsection{Ablation Studies}
We conduct an ablation study to evaluate the contributions of SE-Search's three key components. For a controlled comparison, all models use the same retriever and corpus, are trained for 200 steps, and are evaluated on 500 randomly selected samples from each benchmark. Table~\ref{tab_exp_ablation} reports accuracy on seven benchmarks and also reflects how each component changes search behavior through the final performance.
We compare four configurations.
(1) Search-R1 (baseline);
(2) Search-R1 with MP, augmented with Memory Purification and the memory reward $\mathcal{R}_{\text{mem}}$;
(3) Search-R1 with MP and AP, adding Atomic Query on top of Memory Purification;
(4) Search-R1 with MP, AP, and DR (SE-Search), the full configuration that further includes Dense Reward $\mathcal{R}_{\text{dense}}$; 

\textbf{Each component provides consistent performance gains.} Memory Purification encourages multi-turn search and improves Musique by $5$ points and Bamboogle by $9.1$ points, corresponding to relative increases of $71.43\%$ and $53.85\%$. Atomic Query further strengthens multi-turn search and yields relative improvements of $15.87\%$ on 2Wiki, $35.83\%$ on Musique, and $42.69\%$ on Bamboogle. Overall, the full SE-Search configuration improves performance across seven benchmarks and attains the highest average accuracy.

\begin{table*}[t]
    \centering
    \small
    \resizebox{0.95\linewidth}{!}{
    \begin{tabular}{c|cccccccc}
        \toprule
        \multirow{3}{*}{ Methods } & \multicolumn{3}{c}{Single-Hop QA} & \multicolumn{4}{c}{Multi-Hop QA}       & \multicolumn{1}{l}{} \\
        \cmidrule(lr){2-4} \cmidrule(lr){5-8}
        & NQ    & TriviaQA   & PopQA   & HotpotQA & 2Wiki & Musique & Bamboogle & Avg.                 \\
        \midrule
        SE-Search-3B (EM)   &  0.475 & 0.624  & 0.423  & 0.450 & 0.361 & 0.183 & 0.424 & 0.420 \\ 
        \rowcolor{gray!10} SE-Search-7B (EM)   &  0.467 & 0.629  & 0.461  & 0.438 & 0.355 & 0.195 & 0.488 & 0.433 \\ 
        \rowcolor{gray!20} SE-Search-14B (EM)  &  0.508 & 0.681  & 0.473 & 0.478 & 0.404 & 0.213 & 0.544 & 0.472 \\
        \hline
        SE-Search-3B (F1)   &  0.567 & 0.705  & 0.468  & 0.573 & 0.431 & 0.271 & 0.536 & 0.507 \\ 
        \rowcolor{gray!10} SE-Search-7B (F1)  &  0.556 & 0.705  & 0.504  & 0.559 & 0.422 & 0.290 & 0.601 & 0.519 \\ 
        \rowcolor{gray!20} SE-Search-14B (F1) &  0.598 & 0.751  & 0.514 & 0.602 & 0.478 & 0.294 & 0.631 & 0.553 \\
        \bottomrule
        \end{tabular}
    }
    \vspace{-0.5em}
    \caption{
    Generalization results showing the scaling trend across backbone LLMs of sizes 3B, 7B, and 14B.
    }
    \label{tab_generalization}
\end{table*}

\subsection{Generalization and Scalability}
\paragraph{Search agent follows consistent scaling trend.} We conduct a generalization study across backbone models of different sizes to examine the scaling law~\cite{kaplan2020scalinglaw}. We keep the retriever, corpus, and evaluation protocol unchanged and vary only the backbone scale. Table~\ref{tab_generalization} reports EM and F1 scores using different backbone LLMs, including Qwen2.5-3B, 7B, and 14B, where EM measures exactness, and F1 captures partial overlap. The results are consistent with the scaling law, showing that SE-Search achieves higher average EM and F1 and improves performance on most datasets as the scale of the backbone model increases.

\begin{figure*}[h]
	\centering
	\includegraphics[width=0.98\linewidth]{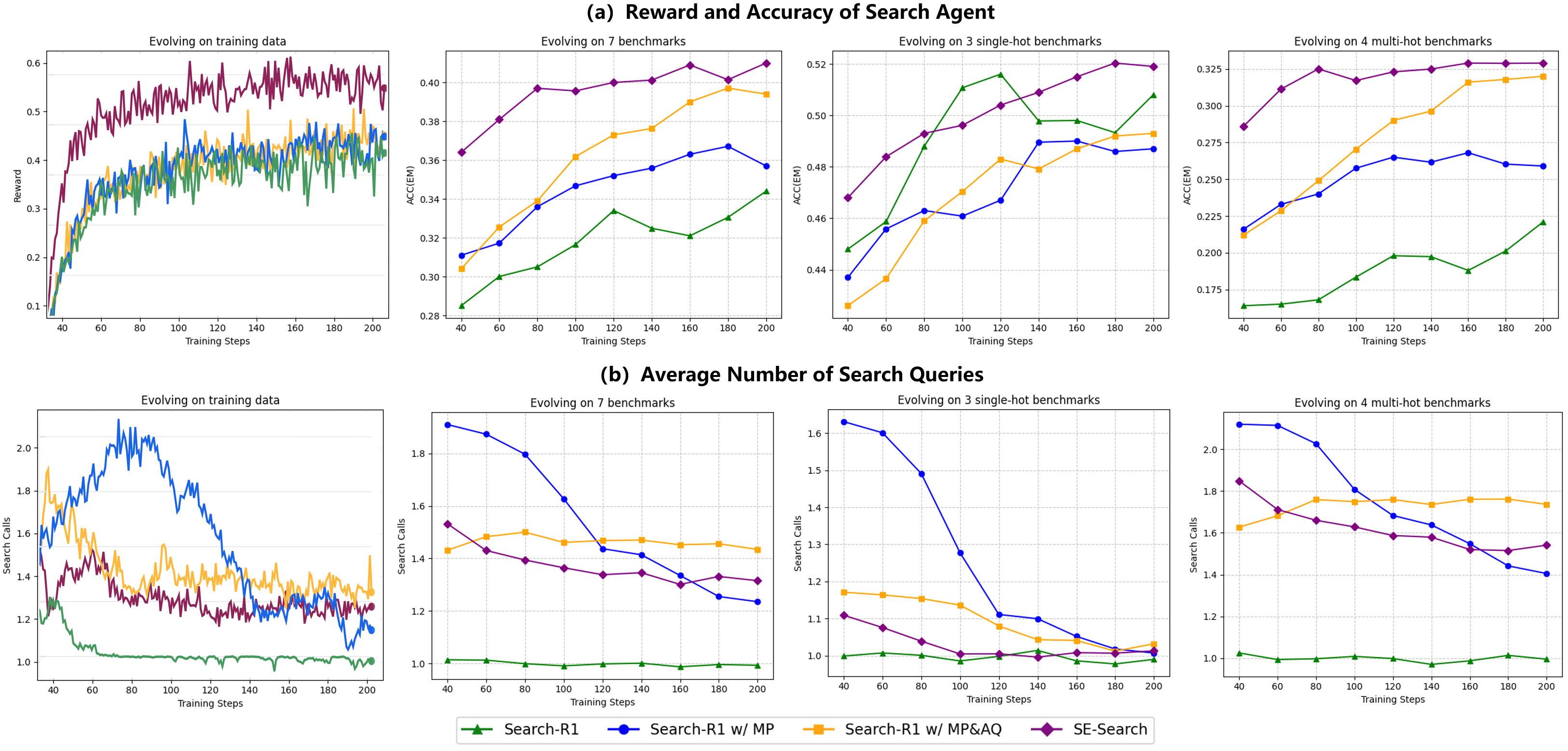}
\caption{SE-Search's evolution of (a) EM accuracy and (b) search calls on training set and benchmarks.}
	\label{fig_searchfreq}
\vspace{-1em}
\end{figure*}

\subsection{Additional Analysis}

\paragraph{Search frequency decreases while accuracy improves}
Figure~\ref{fig_searchfreq} illustrates the evolution of mean accuracy and the mean number of search calls on the training data and evaluation benchmarks. The accuracy of SE-Search increases from $0.36$ to approximately $0.41$, corresponding to an improvement of about $14\%$, while the average number of search calls decreases from $1.53$ to $1.32$, corresponding to a reduction of about $14\%$. We observe that Memory Purification tends to recall a larger set of documents to extract useful information during the early stages of RL. In contrast, Atomic Query generates diverse and high-quality search queries that retrieve fewer but more relevant documents from search engines in a more efficient manner.

\paragraph{SE-Search calls the search engine more often on complex questions.}
Figure~\ref{fig_searchfreq}(b) compares search frequency across training steps and question types, where multi-hop questions represent higher compositional complexity than single-hop ones. Compared with Search-R1, which relies on a fixed single search call, SE-Search issues an average of approximately $1.54$ search calls on multi-hop questions while remaining close to $1.0$ on single-hop benchmarks. This pattern suggests that SE-Search adjusts tool use to question complexity, allocating more retrieval to gather evidence for multi-hop reasoning while avoiding unnecessary calls on simpler questions.

\begin{figure*}[h]
	\centering
	\includegraphics[width=0.98\linewidth]{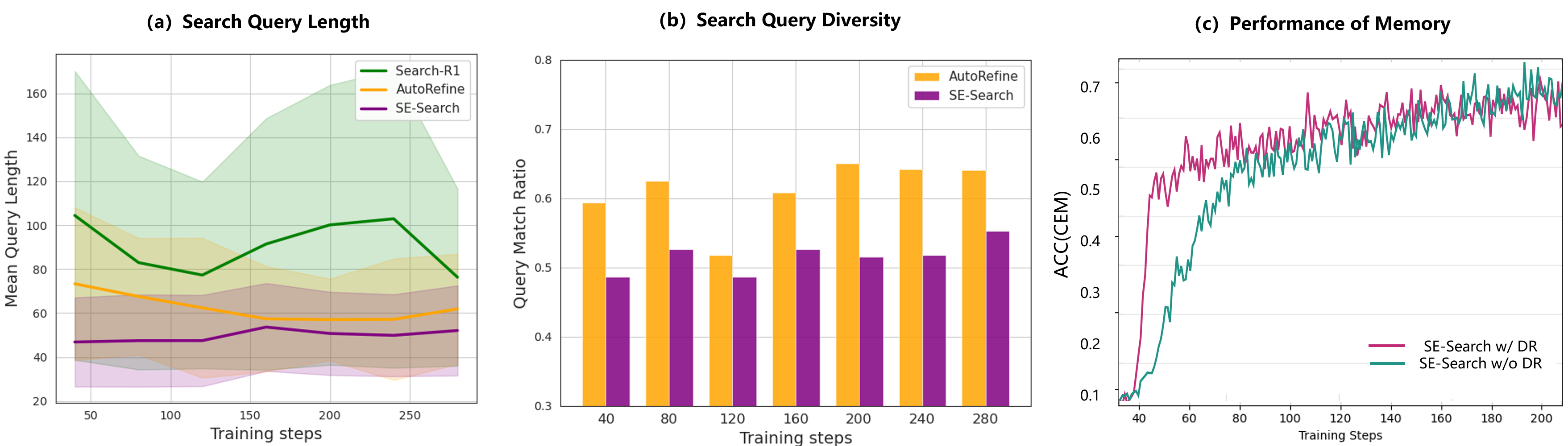}
	\caption{Statistics for (a) search query lengths, (b) search query diversity, (c) accuracy of memory contents.}
	\label{fig_lengthanddiversity}
    \vspace{-1.em}
\end{figure*}
 
\paragraph{SE-Search produces more atomic and diverse queries.}
Prior work emphasizes higher search frequency while largely overlooking query quality. Figure~\ref{fig_lengthanddiversity}(a) reports the average character length and variance of search queries for three methods, including SE-Search at approximately $50$, AutoRefine at approximately $65$, and Search-R1 at approximately $90$. These results suggest that shorter and more focused queries are more effective at retrieving relevant and high-quality documents. Figure~\ref{fig_lengthanddiversity}(b) further compares query diversity using the similarity ratio defined in Eq.~\ref{ratio}. SE-Search achieves a similarity ratio of about $0.5$, whereas AutoRefine reaches a higher value of about $0.6$. Higher query similarity indicates greater redundancy in retrieved results, which may introduce repetitive or biased information and degrade the contextual quality provided to the LLM.

\paragraph{Memory Purification stores key information for answers.}
We evaluate the information stored in memory using CEM, which measures whether the memory contains evidence related to the ground-truth answer. Figure~\ref{fig_lengthanddiversity}(c) shows that the mean CEM of the memory reaches approximately $0.65$ at the final stage of training. This result demonstrates that the agent successfully extracts useful information from retrieved documents and stores key facts in memory, rather than retaining noisy or irrelevant details. Consistent with the training objective, the blue line in Figure~\ref{fig_searchfreq}(b) indicates that the agent issues more search actions during early training to purify memory with important information and filter noise, which leads to higher memory rewards and more reliable memory contents.

\paragraph{Dense Rewards help convergence.}
We further examine the effect of dense rewards, which introduce a format reward and replace the EM-based outcome reward with an F1-based reward. Figure~\ref{fig_lengthanddiversity}(c) shows that dense rewards improve the accuracy of memory content during the early stages of training, indicating earlier learning of how to record answer-relevant evidence. In particular, the format reward discourages invalid outputs and reduces unproductive exploration trajectories, thereby accelerating convergence with fewer RL exploration steps. Moreover, the F1-based outcome reward provides graded feedback even when the answer is not an exact match, which yields denser learning signals than EM and stabilizes training. Consistent with this, Figure~\ref{fig_searchfreq}(a) shows that SE-Search with dense rewards attains higher reward values than SE-Search without dense rewards, and this improvement is mainly attributed to the F1-based outcome reward.

\section{Related Works}

\subsection{Retrieval-Augmented Generation}
Retrieval-Augmented Generation (RAG) improves generative models by grounding outputs in external knowledge~\cite{lewis2020rag}. A typical RAG system consists of a retriever and a generator, and surveys summarize four common paradigms~\cite{gao2023ragsurvey}, including Sequential RAG, Branching RAG, Conditional RAG, and Loop RAG. Conventional RAG often uses a fixed retrieve then generate workflow and does not let the model decide when to search during reasoning. SE-Search instead targets autonomous search behavior with explicit memory, exploration, and behavioral constraints.

\subsection{Search Agent}
Search agents extend RAG by invoking a search tool within a multi-step reasoning process, as in Search-o1~\cite{li2025search-o1}. Related systems such as WebThinker~\cite{li2025webthinker} and WebDancer~\cite{wu2025webdancer} also support autonomous retrieval during reasoning. Other methods focus on training strategies that improve search behavior, including Search-R1~\cite{jin2025searchr1}, ReSearch~\cite{chen2025ReSearch}, AutoRefine~\cite{shi2025autorefine}, and WebSailor~\cite{li2025websailor}. A common limitation is that retrieved context often contains noise that accumulates across steps and weakens later reasoning. SE-Search addresses this issue with Think-Search-Memorize and Atomic Query to improve how evidence is selected and retained.

\subsection{Reinforcement Learning}
Reinforcement learning (RL) provides a general framework for improving LLM behavior. Common approaches include proximal policy optimization (PPO)~\cite{schulman2017ppo}, direct preference optimization (DPO)~\cite{rafailov2023dpo}, and group relative policy optimization (GRPO)~\cite{shao2024deepseekmath}. Extensions such as dynamic sampling policy optimization (DAPO)~\cite{yu2025dapo} and group sequence policy optimization (GSPO)~\cite{zheng2025gspo} further improve training stability and efficiency. GRPO uses group-based normalization and avoids a separate critic. In this work, we design a fine-grained reward function that adapts GRPO to search agent optimization.

\section{Conclusions}
In this paper, we propose SE-Search, a self-evolving search agent that enhances the self-memory, self-exploration, and self-discipline capabilities of LLMs through three evolution directions, including memory purification, atomic query, and dense rewards. SE-Search addresses limitations in prior work that overlook the impact of irrelevant and noisy documents and that lack fine-grained feedback during RL. Extensive experiments on both single-hop and multi-hop question answering benchmarks demonstrate that SE-Search consistently outperforms existing methods.

\section{Limitations}
Although SE-Search outperforms existing methods on single-hop and multi-hop QA benchmarks, it has several limitations. First, SE-Search does not use live web search; the retrieval corpus is static and therefore does not contain real-time information from the internet. Second, highly complex tasks, such as BrowseComp~\cite{wei2025browsecomp}, remain underexplored. Third, the method requires manual selection and tuning of several hyperparameters for dense rewards. Finally, SE-Search supports only a single tool and lacks additional capabilities (e.g., page browsing and code execution).

\bibliography{latex/custom}

\appendix
\section{Appendix}

\subsection{Training and Evaluation Datasets}

Following Search-R1 and other RL-based methods, we merged NQ and HotpotQA to form the training dataset, comprising 169,615 samples. The evaluation set comprises 51,713 samples drawn from the test set of four benchmarks(NQ, TriviaQA, PopQA, Bamboogle) and the development sets of three benchmarks (HotpotQA, 2Wiki, and Musique).

\subsection{GRPO Hyperparameters}

We list the key hyperparameters used with the VeRL framework \cite{sheng2025hybridflow}. In Table~\ref{tab_hyperp}, we adopt the same data, actor, and rollout configuration as Search-R1 and AutoRefine. For the dense reward design, the atomic query reward is computed for query lengths in the range from $10$ to $120$ and uses a similarity threshold of $0.3$. The decay step is set to $150$, which is half of the total training steps.

\begin{table}[ht]
\centering
\caption{Hyperparameters in RL.}
\label{tab_hyperp}
\begin{adjustbox}{width=0.45\textwidth}
\begin{tabular}{ccc}
\toprule
\textbf{Module}  & \textbf{Hyper‑parameter} & \textbf{Value} \\ \midrule
Data & Max Documents Length     & 512 \\
& Total Training Steps & 320 \\
& Max Response Length     & 2048 \\
& Retriever Topk & 3 \\
\midrule
Actor & Training Batch Size      & 256  \\
& Micro Training Batch Size      & 64  \\
& Learning Rate      & $1\times10^{-6}$ \\
& KL Coefficient $\beta$ & 0.001 \\
& Clip Ratio $\epsilon$ & 0.2 \\
\midrule
Rollout & Max Search Actions     & 5 \\
& Group Size $G$ & 10\\
& Temperature     & 1.0 \\
& Top P     & 0.95 \\
\midrule
Reward & $L_{min}$     & 10 \\
& $L_{max}$ & 120 \\
& Similarity Threshold $T$ & 0.3 \\
& $t_{decay}$ & 150 \\
& Reward weights $\alpha$ & 0.1 \\
& Reward weights $\gamma$ & 0.01 \\
\bottomrule
\end{tabular}
\end{adjustbox}
\end{table}

\subsection{More Experiments Results}
\begin{figure*}[h]
	\centering
	\includegraphics[width=0.98\linewidth]{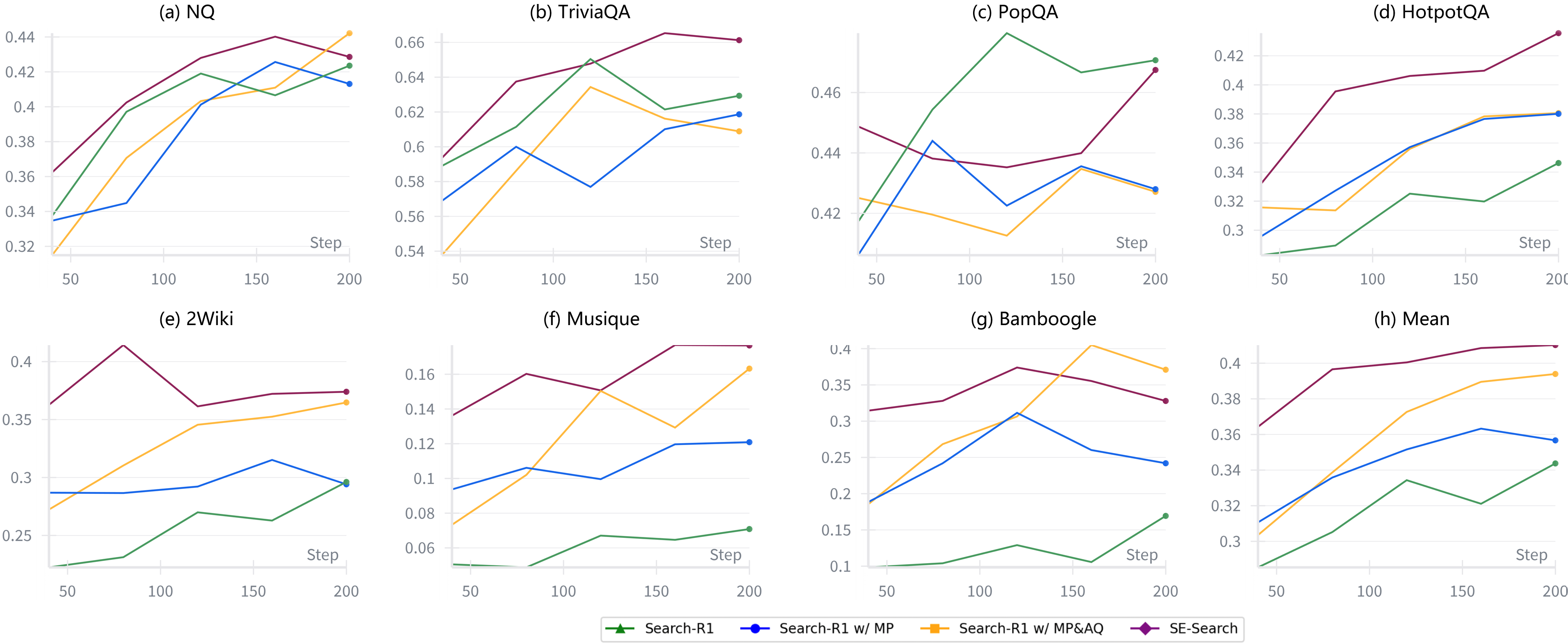}
	\caption{SE-Search's evolution of EM accuracy across seven benchmarks.}
	\label{fig_em7}
    \vspace{-1.em}
\end{figure*}

To provide a more comprehensive characterization of SE-Search, Figure \ref{fig_em7} shows its training dynamics, plotted as average accuracy, across seven benchmarks: Natural Questions (NQ), TriviaQA, PopQA, Bamboogle, HotpotQA, 2Wiki, and Musique.

\subsection{Case Studies}
\label{app_case_study}

Table~\ref{tab_atomic_query} lists the atomic queries derived from a complex user question, illustrating the self-planning and self-exploration of SE-Search. Table~\ref{tab_case1} and Table~\ref{tab_case2} show the memory contents and full trajectories generated by SE-Search for two randomly selected examples from Musique and HotpotQA. These results demonstrate SE-Search's self-memory and self-discipline.

\begin{table*}[p]
    \caption{The list of atomic search queries illustrates self-planning and self-exploration.}
    \vspace{-.5em}
    \label{tab_atomic_query}
    \small
    \resizebox{\linewidth}{!}{
    \begin{tabular}{p{0.98\linewidth}}
    \toprule
    \textbf{Question} Which film has the director died earlier, Kati Patang or A Thief In The Dark? \\
    \textbf{Ground Truth} A Thief In The Dark \\
    \hline
    \textbf{SE-Search}
    \\
    1. \textcolor{green}{\token{search}} Director of Kati Patang  \textcolor{green}{\token{/search}}
    \\
     \textcolor{purple}{\token{memory}} ...  the director of Kati Patang is Shakti Samanta. \textcolor{purple}{\token{/memory}}
    \\
    2. \textcolor{green}{\token{search}}  Director of A Thief In The Dark  \textcolor{green}{\token{/search}}
     \textcolor{purple}{\token{memory}} ... the director of A Thief In The Dark is Donald W. Thompson. \textcolor{purple}{\token{/memory}}
    \\
    \textcolor{blue}{\token{think}} I need to determine which director died earlier, Shakti Samanta or Donald W. Thompson. \textcolor{blue}{\token{/think}} \\
    3. \textcolor{green}{\token{search}} Death date of Shakti Samanta  \textcolor{green}{\token{/search}}
    \textcolor{purple}{\token{memory}} ... Shakti Samanta died on 9 April 2009. \textcolor{purple}{\token{/memory}}
    \\
    4. \textcolor{green}{\token{search}} Death date of Donald W. Thompson  \textcolor{green}{\token{/search}}
    \textcolor{purple}{\token{memory}} ... Thompson died on February 19, 2011. \textcolor{purple}{\token{/memory}} \\
    \textcolor{red}{\token{answer}} A Thief In The Dark \textcolor{red}{\token{/answer}}.
    \\
    \bottomrule
    \end{tabular}
    }
\end{table*}

\begin{table*}[p]
    \caption{Case Studies illustrating self-memory and self-discipline. Key evidence is highlighted \colorEvidence{Blue}.}
    \vspace{-.5em}
    \label{tab_case1}
    \small
    \resizebox{\linewidth}{!}{
    \begin{tabular}{p{0.98\linewidth}} 
    \toprule
    \textbf{Case 1} Who is the spouse of the actor who played Scarlett in Gone with the Wind? \\
    \textbf{Ground Truth} Laurence Olivier \\
    \hline
   
    \textbf{SE-Search}
    \\
    To answer the question "Who is the spouse of the actor who played Scarlett in Gone with the Wind?", I'll follow these steps.
    \\
    1. \textcolor{blue}{\token{think}}I need to identify the actor who played Scarlett in "Gone with the Wind". 
     \textcolor{blue}{\token{/think}}
    \\
    2. \textcolor{green}{\token{search}} Actor who played Scarlett in Gone with the Wind \textcolor{green}{\token{/search}}
    \\
    \token{documents}Doc 1(Title "Gone with the Wind (film)") \colorEvidence{The leading roles are played by Vivien Leigh (Scarlett), Clark Gable (Rhett), Leslie Howard (Ashley), and Olivia de Havilland (Melanie).} ... Doc 3(Title "Vivien Leigh") Vivien Leigh \colorEvidence{Vivien Leigh} (; born Vivian Mary Hartley, and styled as Lady Olivier after 1947; 5 November 1913--8 July 1967) was an English stage and film actress. \colorEvidence{She won two Academy Awards for Best Actress, for her iconic performances as Scarlett O'Hara in ""Gone with the Wind"" (1939)} \textcolor{orange}{\token{/documents}} 
    \\
    3. \textcolor{purple}{\token{memory}} The actor who played Scarlett in "Gone with the Wind" is Vivien Leigh.\textcolor{purple}{\token{/memory}}
    \\
    4. \textcolor{blue}{\token{think}}Now I need to find who Vivien Leigh is married to.  \textcolor{blue}{\token{/think}}
    \\
    5. \textcolor{green}{\token{search}} Vivien Leigh husband \textcolor{green}{\token{/search}}
    \\
    \textcolor{orange}{\token{documents}} Doc 1(Title "Vivien Leigh") ..., \colorEvidence{the public strongly identified Leigh with her second husband, Laurence Olivier}, who was her spouse from 1940 to 1960. Leigh and Olivier starred together in many stage productions, with Olivier often directing, and in three films. Her first husband Leigh Holman also spent considerable ... \textcolor{orange}{\token{/documents}}
    \\
    6. \textcolor{purple}{\token{memory}} Vivien Leigh is married to Laurence Olivier. \textcolor{purple}{\token{/memory}}
    \\
    7. \textcolor{red}{\token{answer}} Laurence Olivier \textcolor{red}{\token{/answer}}.\\
    \bottomrule
    \end{tabular}
    }
\end{table*}

\begin{table*}[p]
    \caption{Case Studies illustrating self-memory and self-discipline. Key evidence is highlighted \colorEvidence{Blue}.}
    \vspace{-.5em}
    \label{tab_case2}
    \resizebox{\linewidth}{!}{
    \begin{tabular}{p{0.98\linewidth}} 
    \toprule
    \textbf{Case 2} How many ethnic minorities were looked at differently in the city where the headquarters of the only group larger than Vilaiyaadu Mankatha's record label is located?\\
    \textbf{Ground Truth} two \\
    \hline
    
    \textbf{SE-Search}
    \\
    To answer the question "How many ethnic minorities were looked at differently in the city where the headquarters of the only group larger than Vilaiyaadu Mankatha's record label is located?", let's follow these steps.
    \\
    1. \textcolor{blue}{\token{think}}First, I need to find out which group larger than Vilaiyaadu Mankatha's record label is located in the headquarters city.  \textcolor{blue}{\token{/think}}
    \\
    2. \textcolor{green}{\token{search}} Which group larger than Vilaiyaadu Mankatha's record label is located in the headquarters city? \textcolor{green}{\token{/search}} 
    \\
    \textcolor{orange}{\token{documents}} Doc 1(Title "Tharangini Records") Primarily established as the first recording and distributing music label in Kerala for Malayalam films... \colorEvidence{ Tharangini Records (currently Tharangni) is the music company} founded by noted Indian Playback singer K. J. Yesudas in 1980 at Thiruvananthapuram, Kerala. It became the first recording centre ...\token{/documents}
    \\
    3. \textcolor{purple}{\token{memory}}  ..., the group larger than Vilaiyaadu Mankatha's record label is Tharangini Records. \textcolor{purple}{\token{/memory}}
    \\
    4. \textcolor{green}{\token{search}} What city is the headquarters of Tharangini Records? \textcolor{green}{\token{/search}}
    \\
    \textcolor{orange}{\token{documents}} Doc 1(Title "Tharangini Records") Tharangini Records Tharangini Records (currently Tharangni) is the music company founded by noted Indian Playback singer K. J. Yesudas in 1980 ... \colorEvidence{The office and studio were moved to Abhiramapuram, Chennai, Tamil Nadu licensed under Dr. K. J. Yesudas Audio and Visual Corporation Ltd.}) ...\token{/documents}
    \\
    5. \textcolor{purple}{\token{memory}} ..., the headquarters of Tharangini Records is located in Chennai, Tamil Nadu.  \textcolor{purple}{\token{/memory}}
    \\
    6. \textcolor{green}{\token{search}}  How many ethnic minorities were looked at differently in the city of Chennai, Tamil Nadu? \textcolor{green}{\token{/search}}
    \\
    \textcolor{orange}{\token{documents}} Doc 1(Title "Demographics of Chennai") ... however, the population growth rate was slow. Hindus form the majority of Chennai's population \colorEvidence{but the city also has substantial Muslim and Christian minorities.} \textcolor{orange}{\token{/documents}}
    \\
    7. \textcolor{purple}{\token{memory}} ..., the city of Chennai, Tamil Nadu has a substantial Muslim and Christian minority. \textcolor{purple}{\token{/memory}}
    \\
    8. \textcolor{red}{\token{answer}} Two \textcolor{red}{\token{/answer}}.\\
    \bottomrule
    \end{tabular}
    }
\end{table*}

\end{document}